# Relevant Explanations: Allowing Disjunctive Assignments

Solomon Eyal Shimony
Math. and Computer Science Department
Ben Gurion University of the Negev
P.O. Box 653, 84105 Beer-Sheva, Israel
shimony@bengus.bitnet

## Abstract

Relevance–based explanation is a scheme in which partial assignments to Bayesian belief network variables are explanations (abductive conclusions). We allow variables to remain unassigned in explanations as long as they are irrelevant to the explanation, where irrelevance is defined in terms of statistical independence. When multiple-valued variables exist in the system, especially when subsets of values correspond to natural types of events, the overspecification problem, alleviated by independence-based explanation, resurfaces. As a solution to that, as well as for addressing the question of explanation specificity, it is desirable to collapse such a subset of values into a single value on the fly. The equivalent method, which is adopted here, is to generalize the notion of assignments to allow disjunctive assignments.

We proceed to define generalized independence based explanations as maximum posterior probability independence based generalized assignments (GIB-MAPs). GIB assignments are shown to have certain properties that ease the design of algorithms for computing GIB-MAPs. One such algorithm is discussed here, as well as suggestions for how other algorithms may be adapted to compute GIB-MAPs. GIB-MAP explanations still suffer from instability, a problem which may be addressed using "approximate" conditional independence as a condition for irrelevance.

## 1 INTRODUCTION

Explanation, finding causes for observed facts (or evidence), is frequently encountered within Artificial Intelligence. Research and applications exist in natural language understanding [10, 1, 19], automated medical diagnosis [5, 14, 13], vision and image processing [7, 6], finding commonsense explanations, and other fields [15]. In general, finding an explanation is characterized as follows: Given world knowledge in the form of (causal) rules, and observed facts (a formula), determine what needs to be assumed in order to *predict* the evidence.[1]

One would like to find an explanation that is "optimal" in some sense. Systems that perform explanation tasks need to provide criteria for optimality. In related papers [18, 16, 17], we have argued that plausibility, the power of predicting the observed facts, and relevance, are important criteria. We assume a framework that has causality as a primitive notion, and uses probabilities as the uncertainty formalism. World knowledge in this framework can be represented in the form of Bayesian belief networks. Random variables in the network (also referred to as *nodes* throughout) are assumed to represent the occurrence of real-world events. For simplicity, we assume that the nodes are discrete random variables. Evidence is an assignment of values to some of the nodes in the network, while an explanation is another such assignment that obeys the plausibility, predictiveness, and relevance criteria. Note that an assignment here is treated as a sample-space event, and as such has a probability. For example, if we have a random variable die-throw, then the assignment die-throw=3 is the event where the die turns up with a 3, and has a probability of $\frac{1}{6}$, assuming a fair 6 sided die.

With the above assumptions, optimizing plausibility and predictiveness means maximizing the posterior probability of the explanation (or assignment). If we ignore relevance, then just finding the MAP (Maximum A-Posteriori) assignment to the network is sufficient. The necessity for relevance was shown by example in [11], by noting that assigning values to irrelevant variables leads to anomalous abductive conclusions. It was suggested that only nodes that are ancestors of some evidence node ("evidentially supported") be assigned.

In [18, 16, 17], we presented a variant of the example

---

[1] Thus, by "explanation" we mean "abduction", "abductive reasoning", or "diagnosis", and "an explanation" is "abductive conclusions". We do *not* intend to imply that such explanations are for human consumption.

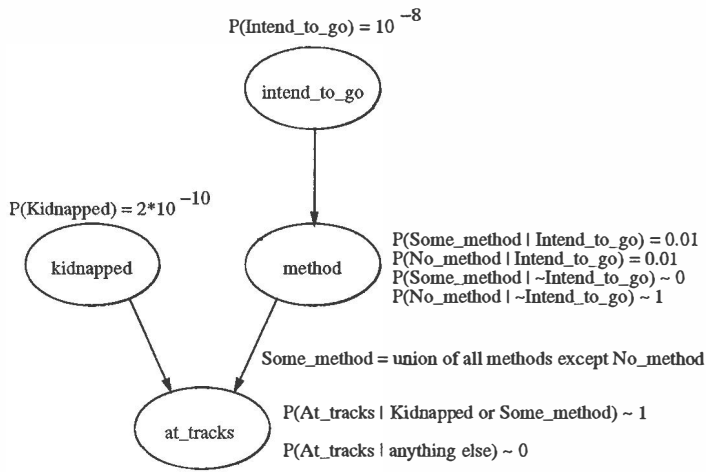

Figure 1: Train Tracks Example Network

(the vacation plan problem, omitted here for lack of space), and called that form of anomaly the "overspecification problem". We noted that the evidential support criterion still considered too many nodes as relevant. We then defined explanations as *partial* MAPs, i.e. assignments of maximum probability where irrelevant nodes are left unassigned. The evidence nodes are always considered to be relevant. Nodes are considered irrelevant if they are not ancestors of some evidence node (evidentially supported), or if a certain statistical independence criterion holds for all of their relevant descendants (the Independence-Based (IB) condition):

**Definition 1** *The IB condition holds at node $v$ w.r.t. an assignment $\mathcal{A}$ iff $v$ is independent of the ancestors of $v$ that are not assigned by $\mathcal{A}$, given the values assigned by $\mathcal{A}$ to (the rest of) the ancestors.*

By using statistical independence in this way to define irrelevance, the overspecification problem was partially alleviated[2]. Several problems remained in the solution: "almost" independent cases (which may be overcome by using $\delta$-IB MAP explanations [3] [17]), and incapability of providing disjunctive explanations, even when the representation is favorable. Consider the following example (figure 1):

Jack is found at the train tracks (our evidence, which needs explaining). Suppose that there are two explanations for his being there: getting there of his own accord, or being kidnapped. For getting there intention-

---

[2] The idea of using independence (in Bayesian belief networks) of a *particular assignment* to a set of variables (rather than all assignments to the variables) is similar to that of Bayes multinets [8].

[3] $\delta$-IB MAP explanations are the same as IB MAP explanations, except that in the IB condition, "independent" is replaced by "independent within a factor of $1 - \delta$", i.e. the ratio between the maximal and minimal conditional probabilities over all possible assignments to the ancestors of $v$ is greater than $1 - \delta$.

---

ally, Jack may have used any one of 99 different methods (such as walking, taking a bus, etc.), all equally likely given that Jack intended to get to the tracks, for the sake of this example. The method variable is represented by a node with 100 possible values, one for each method, and one for not going by any method (for the case where he did not intend to go, or could not go for other reasons). Assume that the prior probability of getting kidnapped is 50 times less than that of intending to go.

Since the IB condition does not hold at the at-tracks node given that kidnapping did not occur (nor does the $\delta$-IB condition hold, for any reasonable $\delta$), the system would prefer the kidnapping explanation. Intuitively, we should prefer the intend-to-go explanation, and should just ignore the method node, or state that Jack went to the train-tracks by some (undetermined) method. Even if we use a weighted abduction system, such as [10, 2], the problem still remains. We noted in [18] that if we allowed the system to collapse all the methods into a single method, or equivalently allowed *disjunctive* assignments, the problem would go away. Actually, the method of selecting nodes with high posterior probabilities to be part of the explanation also happens to give the right answer, but we have already shown in [18, 17], that the posterior node probabilities scheme is undesirable for other reasons (possible inconsistencies and irrelevant explanations).

Disjunctive assignments in explanations are also useful for handling cases where there are multiple-valued variables, such that sets of values correspond to natural types of events in a taxonomic hierarchy. In this case we might want to aggregate the values into a single value on the fly, if the need arises. We formally develop disjunctive assignments (which we also call generalized assignments) in section 2, and use them to define generalized IB (GIB) assignments and explanations. GIB assignments are shown to have locality properties similar to those that hold for IB assignments. Section 3 discusses an algorithm for implementing GIB explanation, based on the locality properties. Section 4 evaluates GIB explanation, and suggests how to extend the formalism to handle $\delta$ independence.

## 2 GIB EXPLANATION

When forming an explanation, we need to decide whether certain events are part of the explanation. Several (perhaps even "most") AI programs use a taxonomic hierarchy for representing event types (as well as other kinds of object types). One interesting question is that of the specificity of the explanation: should we prefer an event higher up the hierarchy (more general), or lower down (more specific)?

A solution proposed by Goldman and Charniak for the WIMP natural language understanding program [1, 9] allows aggregation of node values into a single value. The kind of specificity that this scheme handles is specificity of event description w.r.t. some hierarchical knowledge base of events. For example, suppose that one event type is "shopping", and that there are events lower down in the hierarchy, "supermarket shopping", "liquor-store shopping", etc. that are subtypes of "shopping". In the belief network representation, a multiple valued node consisting of all possible events is used. Posterior probabilities are computed. If the probabilities of the individual subtypes of shopping events is low, one may still aggregate all these into a single value that corresponds to "shopping", and if that has a high probability, a decision on the shopping explanation can be made. In this example, the system selects a less specific explanation (less specific, at least, than a particular subtype of shopping), in order to get a high probability explanation.

This scheme works only if the taxonomic hierarchy is a strict hierarchy, i.e. each object has only one parent and there are no "negative" links. We will assume that this is indeed the case, as is done in WIMP. This means that the is-a hierarchy does not have multiple inheritance. The implication of this is that the number of possible aggregated values for a node with $n$ possible values is at most $2n$.

We have seen in the introduction how allowing aggregation of node values can help us alleviate the over-specification problem. Rather than actually aggregating values into a single value, we elect to generalize assignments. Assignments can now assign a disjunction of values to a node or variable. The result will be the same as when aggregating node values into a single value. We do not want to allow any old disjunction to be assigned, however. The disjunctions assigned should correspond to concepts, or to different events in our hierarchy of event types. The most general event is the "anything happens" event, which corresponds to the disjunction of all the values of a node. Assigning the "anything happens" disjunction to a node, is exactly equivalent to leaving it unassigned. Thus, we see that allowing the assignment of disjunctions to nodes in explanations is a generalization of independence-based explanations.

We remain with the question: when do we allow a particular disjunction to be assigned to a node in a proposed explanation? The answer to this question is not at all obvious. For example, if we allowed any disjunction corresponding to a concept to be used every time, then all explanations will assign the most general disjunction (a disjunction of all the node's values) to each node. Essentially, this is equivalent to leaving all non-evidence nodes unassigned, which gives us the highest probability assignment. This result is, however, an undesirable trivial explanation, that is completely independent of our knowledge base.

Instead, we propose the following criteria: first, the disjunction has to correspond to a pre-existing concept. The reason for this assumption is that we want an explanation to consist only of natural events and concepts. This is equivalent to assuming that a set of allowable disjunctions is provided to the system. Second, we only assign a disjunction if the probability of the descendent nodes is statistically independent of which value (from the disjunction) we condition on.

To get a picture of where this is leading us, consider the special case where the only higher level concept is the "any event" concept. In this case, allowing the assigning of disjunctions under the above constraints is exactly equivalent to independence based assignments. That is because the only allowed disjunctions are those with a single value, or those with all the values of a node. The second constraint forces us to assign the disjunction only if independence occurs, exactly as in the case of independence-based assignments.

We will ignore in this paper the representation issue, and just assume that for each (multi-valued) node, a set of all permissible disjunctive assignments is given, in some form. Thus, for each node $v$ in the belief network, with a domain $D_v$, the set of permissible disjunctions $M_v$ is given, where $M_v \subseteq 2^{D_v}$, as well as the set of all conditional probabilities of each permissible disjunctive assignment to $v$ given the parents of $v$. In what follows, we will usually omit referring to $M_v$, assuming its presence implicitly.

One may argue that we do not need to introduce the first constraint and $M_v$ at all. We could allow any disjunction, as long as the second constraint, that conditional independence hold, is obeyed. In fact, this seems equivalent to an argument of the following form: we (as intelligent agents) construct our concepts from empirical data. Therefore, if (conditional) independence occurs, i.e. it does not matter which of a set of values is assigned, we are justified in creating a new concept that corresponds to that set of values. The latter argument seems reasonable, but this issue is beyond the scope of this paper. Suffice it to say that our definitions require the existence of the set of allowable disjunctions $M_v$, but if we decide that it is not needed, we can just set $M_v = 2^{D_v}$ for every variable in the network, thereby voiding the first constraint.

## 2.1 GIB EXPLANATION: DEFINITION

We begin by formally defining assignments and disjunctive (or generalized) assignment. An assignment $\mathcal{A}$ to a set of variables $V$, each variable $v \in V$ having domain $D_v$, is a set of pairs $(v, d)$, where $v \in V$ and $d \in D_v$. If $(v, d) \in \mathcal{A}$ we say that $\mathcal{A}$ assigns variable $v$ the value $d$. We sometimes write $v = d$ instead of $(v, d)$ in an assignment. In our example, we might have an assignment:

$$\mathcal{Q} = \{\text{at-tracks} = T, \text{method} = \text{walk}\}$$

$\mathcal{A}$ is *complete* w.r.t. $V$ if for every $v \in V$ there is a pair $(v, d) \in \mathcal{A}$ for some $d$, i.e. it assigns values to all of the variables. We call an assignment *partial* if it is not necessarily complete.

We define span($\mathcal{A}$) to be the set of variables assigned by $\mathcal{A}$, i.e. span($\mathcal{A}$) = $\{v | \exists d\ (v, d) \in \mathcal{A} \wedge d \in D_v\}$. For example, span($\mathcal{Q}$) = {at-tracks, method}. $\mathcal{A}$ is *consistent* if each variable in the span of the assignment is assigned a unique value, i.e. if $(v, d) \in \mathcal{A}$ and $(v, d') \in \mathcal{A}$ then $d = d'$.

A disjunctive (or generalized) assignment $\mathcal{A}$ to a set of variables $V$ is a set of pairs $(v, D)$ where $v \in V$ and $D \subseteq D_v$. Each variable is assigned a set of values, rather than just a single value. A generalized assignment is also a sample space event, the union of the events comprising its member assignments. In some cases, we use the notation $v = d_1 \vee d_2 \vee ... \vee d_k$ as a variant for $(v, \{d_1, d_2, ..., d_k\})$. In our example, we might have:

$$\mathcal{G} = \{\text{at-tracks} = T, \text{method} = \text{take-taxi} \vee \text{walk}\}$$

For generalized assignment (G-assignment) $\mathcal{A}$, we define span($\mathcal{A}$) to be the set of variables assigned by $\mathcal{A}$, i.e. span($\mathcal{A}$) = $\{v | \exists D\ (v, D) \in \mathcal{A} \wedge D \subseteq D_v\}$. span$^-(\mathcal{A})$, the *proper span* of $\mathcal{A}$, is the set of variables $v$ that are assigned a value-set different from $D_v$. Formally[4]: span$^-(\mathcal{A}) = \{v | \exists D\ (v, D) \in \mathcal{A} \wedge D \subset D_v\}$. A G-assignment is consistent if it assigns a unique, non-empty set to each variable in the span of the assignment, i.e. if $(v, D) \in \mathcal{A}$ and $(v, D') \in \mathcal{A}$ then $D = D' \neq \phi$.

G-assignment $\mathcal{B}$ is *more refined* than G-assignment $\mathcal{A}$ (written $\mathcal{B} \subseteq \mathcal{A}$) iff every value set assigned by $\mathcal{A}$ to each variable is a (non-strict) superset of the value set assigned by $\mathcal{B}$. Formally:

$$\mathcal{B} \subseteq \mathcal{A} \leftrightarrow ((v, D) \in \mathcal{A} \rightarrow \exists D'\ (v, D') \in \mathcal{B} \wedge D' \subseteq D)$$

Likewise, G-assignment $\mathcal{B}$ is *strictly more refined* than G-assignment $\mathcal{A}$ (written $\mathcal{B} \subset \mathcal{A}$) iff every value set assigned by $\mathcal{A}$ to each variable is a (non-strict) superset of the value set assigned by $\mathcal{B}$, except for at least one variable, where the value set is a *strict* superset.

An assignment $\mathcal{A}$ is *included* in a G-assignment $\mathcal{B}$ (written $\mathcal{A} \dot{\in} \mathcal{B}$) if for every node $v$ assigned some value set $D$ by $\mathcal{B}$, the node is assigned a value in $D$ by $\mathcal{A}$. That is, $(v, D) \in \mathcal{B} \rightarrow (\exists d\ d \in D \wedge (v, d) \in \mathcal{A})$. For example, both {at-tracks=T, method=take-taxi} and {at-tracks=T, method=walk} are included in $\mathcal{G}$.

Sometimes we need to refer to an assignment (or G-assignment) to only certain variables, possibly a subset of the span of some assignment. We denote such (partial) assignments with a subscript, the set of nodes in the partial assignment. Thus, for an assignment (or G-assignment) $\mathcal{A}$:

$$\mathcal{A}_S = \{(v, D) | (v, D) \in \mathcal{A} \wedge v \in S\}$$

**Definition 2** *The generalized independence-based condition (GIB condition) holds at node $v$ w.r.t. G-assignment $\mathcal{A}$ iff:*

$$\forall \mathcal{B} \subseteq \mathcal{A}_{\uparrow(v)}\quad P(\mathcal{A}_{\{v\}} | \mathcal{B}_{\uparrow^+(v)}) = P(\mathcal{A}_{\{v\}} | \mathcal{A}_{\uparrow(v)})$$

where $\uparrow(v)$, denotes the parents (direct predecessors) of $v$, and $\uparrow^+(v)$ denotes the transitive closure of parents of $v$ (here and throughout this paper).

Intuitively, the GIB condition holds at $v$ if the conditional probability of $v$ given the G-assignment $\mathcal{A}$ to the parents of $v$ is independent of the way we refine the G-assignment w.r.t. the ancestors of $v$ (i.e. independent of any further evidence coming from above). We proceed to define GIB assignments as assignments where the GIB condition holds at every node. Formally:

**Definition 3** *A generalized assignment $\mathcal{A}_S$ is GIB iff for every node $v \in S$, the GIB condition holds.*

Finally, we define a GIB MAP as the most probable GIB assignment where the evidence nodes are assigned correctly. Formally:

**Definition 4** *A generalized assignment $\mathcal{A}_S$ is a GIB MAP w.r.t. evidence $\mathcal{E}$ iff it is a maximum probability GIB assignment such that $\mathcal{E} \subseteq \mathcal{A}$.*

A GIB explanation is a compact GIB MAP, i.e. a GIB MAP without the pairs $(v, D)$ such that $D = D_v$. Such value-set assignment pairs contribute no information, and are thus excluded from the explanation. Note also that there it is sufficient to maximize the prior probability $\mathcal{A}_S$ (as defined above), rather than the conditional probability $P(\mathcal{A}|\mathcal{E})$, as the evidence is constant for each problem instance, and $P(\mathcal{E}|\mathcal{A}) = 1$. In our train-tracks example, the GIB MAP is $\mathcal{M} =$ {at-tracks=T, method=$m_1 \vee m_2 \vee ... \vee m_{99}$, intend-to-go=T}, where each $m_i$ is one of the methods $\mathcal{M}$ is a GIB assignment because at-tracks is independent of the value of the "kidnapped" node, or the assignment to the method node. It is the GIB-MAP because it is the most probable amongst the GIB assignments that have at-tracks=T, with a prior probability of approximately $10^{-8}$.

### 2.2 PROPERTIES OF GIB ASSIGNMENTS

The independence relations that underlie Bayesian belief networks induce certain locality properties on GIB assignments. These are useful for designing algorithms that compute GIB explanations. We begin by showing that the bounds on the conditional probability of a node can be obtained using the bounds of the conditional probability of local complete assignments, i.e. assignments to the parents (ignoring all the other ancestors):

**Theorem 1** *For positive distributions, the following equations hold:*

$$\min_{\mathcal{B} \subseteq \mathcal{A}_{\uparrow(v)}} P(\mathcal{A}_{\{v\}} | \mathcal{B}_{\uparrow^+(v)}) = \min_{\mathcal{D} \in \mathcal{C}_{\uparrow(v)} \wedge \mathcal{D} \dot{\in} \mathcal{A}_{\uparrow(v)}} P(\mathcal{A}_{\{v\}} | \mathcal{D})$$

---
[4]since assigning $D_v$ to $v$ does not restrict the possible values that $v$ may have, we sometimes would like to say that $v$ is not "properly" assigned in this case.

$$\max_{\mathcal{B} \subseteq \mathcal{A}_{\uparrow(v)}} P(\mathcal{A}_{\{v\}}|\mathcal{B}_{\uparrow^+(v)}) = \max_{\mathcal{D} \in \mathcal{C}_{\uparrow(v)} \wedge \mathcal{D} \dot{\in} \mathcal{A}_{\uparrow(v)}} P(\mathcal{A}_{\{v\}}|\mathcal{D})$$

where $\mathcal{C}_S$ is used to denote the set of all complete assignments to node set $S$ (throughout this paper), and thus $\mathcal{D}$ ranges over all the *complete* assignments to the parents of $v$ that are included in $\mathcal{A}_{\uparrow(v)}$, and $\mathcal{B}$ ranges over all G-assignments that are refinements of $\mathcal{A}_{\uparrow(v)}$. For a proof, see appendix A.

From these bounds (theorem 1), and the definition of the GIB condition (definition 2), it is easy to show that the GIB condition holds at a node if conditional independence holds locally:

**Theorem 2** *For positive distributions, the GIB condition holds at $v$ w.r.t. G-assignment $\mathcal{A}$ iff the following equation holds:*

$$\min_{\mathcal{D} \in \mathcal{C}_{\uparrow(v)} \wedge \mathcal{D} \in \mathcal{A}_{\uparrow(v)}} P(\mathcal{A}_{\{v\}}|\mathcal{D}) = \max_{\mathcal{D} \in \mathcal{C}_{\uparrow(v)} \wedge \mathcal{D} \in \mathcal{A}_{\uparrow(v)}} P(\mathcal{A}_{\{v\}}|\mathcal{D})$$

Thus, checking whether an assignment is GIB is linear in the size of the span of the assignment, and does not depend on the size of the graph. Additionally, if $\mathcal{A}$ is a GIB assignment, then its probability is a simple product, computable in time linear in $|span(\mathcal{A})|$.

**Theorem 3** *Let $\mathcal{A}$ be a GIB assignment to a (positive distribution) Bayesian belief network. $P(\mathcal{A})$, the probability of $\mathcal{A}$, is the product:*

$$P(\mathcal{A}) = \prod_{v \in span(\mathcal{A})} P(\mathcal{A}_{\{v\}}|\mathcal{A}_{\uparrow(v)}) \quad (1)$$

This is an important property, as to compare quality of GIB explanations, we need to know their probability, and this theorem allows us to do so efficiently. A proof outline is discussed in appendix A.

Note that the restriction to positive distributions in the theorems is only needed so as to ensure that all the conditional probabilities referred to (in the theorems and their proofs) are defined. Thus, as long the latter requirement holds, we do not need the restriction to positive distributions.

## 3  GIB-MAP ALGORITHM

An algorithm that uses best-first search is presented in what follows. The search space is that of partial generalized assignments (not only GIB assignments), beginning with the assignment denoting the evidence, and concluding with a GIB assignment of maximum probability given the evidence. The next-state generator selects a node $v$ and generates assignments that are refinements of the current assignment, by refining the assignment to the parents of $v$.

The algorithm is essentially a generalization of the algorithm for finding IB-MAPs [16], achieved by generalizing the concept of hypercubes, on which the IB-MAP algorithm is based, to allow for disjunctive assignments. Generalized hypercubes are generalized assignments that assign permissible disjunctions to a node and its parents.

**Definition 5** *A generalized assignment $\mathcal{A}$ is a generalized hypercube (G-hypercube) based on node $v$ iff $span(\mathcal{A}) = \{v\} \cup \uparrow(v)$, and if $w \in span(\mathcal{A})$ then $\mathcal{A}(w) \in M_w$.*

We essentially assume that $M_w = D_w$, so that the intersection of two value sets (assigned to a variable $v$ by two different G-hypercubes) is always a permissible value set for $v$ in a G-hypercube. The latter requirement may be satisfied by less restrictive assumptions, but this issue is beyond the scope of this paper. We define maximal generalized IB hypercubes, in a manner similar to IB hypercubes in [16].

**Definition 6** *A G-hypercube $\mathcal{A}$ based on $v$ is a GIB hypercube (based on $v$) iff the generalized IB condition holds at $v$ w.r.t. $\mathcal{A}$.*

**Definition 7** *GIB hypercube $\mathcal{A}$ is maximal if it is minimal w.r.t. refinement, i.e. there is no GIB hypercube $\mathcal{B}$ such that $\mathcal{A} \subset \mathcal{B}$.*

In our example, the assignment {at-tracks=T, method=$m_1 \vee m_2 \vee ... \vee m_{99}$, kidnapped=T} is a GIB-hypercube based on the node at-tracks, which is a refinement of the GIB-hypercube {at-tracks=T, method=$m_1 \vee m_2 \vee ... \vee m_{99}$}. The latter is minimally refined, and is thus a maximal GIB-hypercube.

The algorithm is shown in figure 2. The termination condition is that the G-IB condition hold at every node (it is a weaker condition than the IB-condition). The GIB condition holds at every expanded node, so there is no need to check the condition explicitly for every node in the assignment. It is sufficient that all nodes are expanded.

States are partial G-assignments, augmented with a value (approximate probability) and an (integer) index of the node last expanded. The agenda is kept sorted (in a heap) by its approximate probability, which for each state $\mathcal{A}$ is determined by:

$$P_a(\mathcal{A}) = \prod_{v \in S} P(\mathcal{A}_{\{v\}}|\mathcal{A}_{\uparrow(v)})$$

where $S$ is the set of expanded nodes in $\mathcal{A}$. Theorem 3 ensures that $P_a$ is an admissible heuristic evaluation function, as it is correct for GIB assignments (all nodes expanded), and is optimistic for any other assignment in the agenda.

When picking a node, the algorithm selects the unexpanded node with smallest index in the assignment. Node indexing is such that each node has a smaller index than all of its ancestors. Clearly this can be done, as belief networks are DAGs. The ordering is not necessarily unique, and we just pick some such ordering.

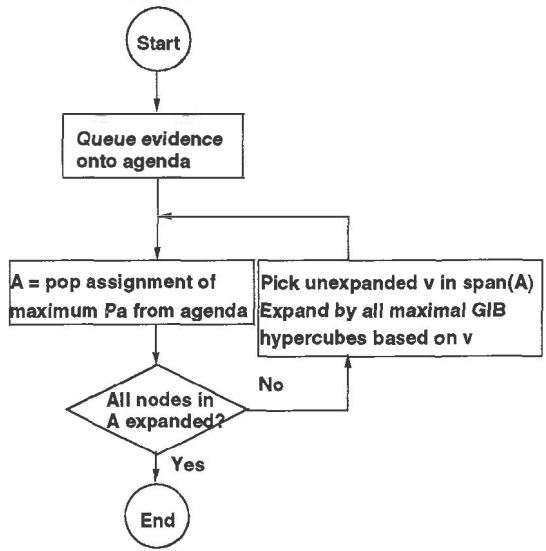

Figure 2: Computing GIB Explanations

For each assignment, save the number of the node $v$ last expanded.

To expand a node $v$, first check if the GIB condition holds at $v$. The condition holds vacuously for root nodes. Otherwise, if there exists a GIB hypercube $\mathcal{H}$ such that $\mathcal{A}_{\uparrow(v)} \subseteq \mathcal{H}$, then the GIB condition holds at $v$. Set last-expanded($\mathcal{A}$) to the index of $v$. If the GIB condition holds at $v$, the we consider it expanded, so evaluate $\mathcal{A}$, and push it back into the agenda.

Otherwise, select all maximal GIB hypercubes $\mathcal{H}$ that are refinements of the assignment $\mathcal{A}$ for $v$ and its parents, i.e. such that $\mathcal{H} \subseteq \mathcal{A}_{\uparrow(v)}$. For each such GIB hypercube, generate one new assignment $\mathcal{B}$ as follows: $\mathcal{B}$ is a (minimally refined) refinement of both $\mathcal{H}$ and $\mathcal{A}$. This is done by looking at the assignment to each node $v$. If the node is assigned by only one of the assignments a value set $D$, then $(v, D)$ is in $\mathcal{B}$. Otherwise (if $(v, D^\mathcal{A}) \in \mathcal{A}$ and $(v, D^\mathcal{H} \in \mathcal{H})$ for some $D^\mathcal{A}$ and $D^\mathcal{H}$), then $(v, D^\mathcal{A} \cap D^\mathcal{H})$ is in $\mathcal{B}$. Evaluate each such $\mathcal{B}$ generated above, and push it into the agenda.

As to the complexity of the algorithm, finding abductive conclusions is known to be NP-hard in the propositional case [3, 18], so that any algorithm may be exponential time in the worst case, as indeed is the case for our algorithm. However, timing experiments made for the very similar IB-MAP algorithm suggest that in practice the running time is reasonable.

## 4 DISCUSSION

We have shown how generalizing assignments to disjunctive assignments, allows us to be more flexible in defining independence, so as to alleviate the overspecification problem when we have multiple-valued variables, in which sets of values stand for natural types of events. We do not think, however, that the overspecification problem is completely overcome by GIB MAP explanation. That is because slightly changing conditional probabilities may cause an overspecified assignment to variables that are still intuitively irrelevant, which may in turn cause the wrong explanation to be preferred.

This instability problem shown above becomes particularly acute if the belief network is constructed using probabilities calculated from real statistical experiments. That can be done either by first constructing the topology of the network and experimenting to fill in the conditional probabilities, or by using a method such as in [4] or as in [12] to get the topology as well as the conditional probabilities directly from the experiments. In either case, even if exact independence exists in the real world, the conditional probabilities computed based on experiments are very unlikely to be *exactly* equal.

The problem of "almost" independent cases, as well as a solution that uses $\delta$ independence, is explored in [18, 17]. It should be possible to apply $\delta$ independence to generalized assignments as well, as follows:

**Definition 8** *The generalized delta independence-based condition ($\delta$-GIB condition) holds at node $v$ w.r.t. G-assignment $\mathcal{A}$ iff*

$$\min_{\mathcal{B} \subseteq \mathcal{A}_{\uparrow(v)}} P(\mathcal{A}_{\{v\}}|\mathcal{B}_{\uparrow+(v)}) \geq (1-\delta) \max_{\mathcal{B} \subseteq \mathcal{A}_{\uparrow(v)}} P(\mathcal{A}_{\{v\}}|\mathcal{B}_{\uparrow+(v)})$$

This is a parametric definition: with $\delta = 0$ (most restrictive), we get the GIB condition. With $\delta = 1$, the condition always holds. The correct value for $\delta$ is not obvious, and it may be desirable to choose its value on a per-node basis. That may be done, if the distributions are obtained from empirical data, by estimating the sampling error bounds. Alternately, we may wish to bias $\delta$ based on prior probabilities of the parents of $v$.

As for properties of GIB assignments, we believe that a variant of theorem 2, that allows local checking of the $\delta$-GIB condition, holds due to theorem 1. It is clear that theorem 3 does not hold for $\delta$-GIB assignments, however.

## 5 SUMMARY

We have shown that generalizing irrelevance-base explanations to allow a limited assignment of disjunctions (GIB assignments) further alleviates the overspecification problem. We get the added bonus that the disjunction allows us to choose a less specific event (in a particular node) as long as it is irrelevant which event subtype occurred. GIB assignments were shown to have certain locality properties.

Based on the locality properties, a best-first search algorithm for finding GIB MAPs was easy to define along the lines of an earlier algorithm, that for computing IB MAPs. It should be possible to show that the problem is naturally reducible to linear programming (with a 0-1 solution requirement), as was done for IB MAPs in [18], which provides another possible algorithm for computing GIB MAPs. It would also be interesting to prove the locality property for $\delta$-GIB assignments, and propose an algorithm for computing them, perhaps similar to the algorithm for $\delta$-IB MAP computation, which uses bounds on the probability of a $\delta$-IB assignment, rather than its exact probability.

Another issue for future research is the following: The fact that we are using disjunctive assignments rather than single value assignments may allow us to extend IB explanations to handle continuous random variables as well. Events would be ranges of such random variables where over which conditional independence holds (i.e. intervals where conditional density function is constant).

## A  PROOFS FOR THEOREMS

**Theorem 1** *For positive distributions, the following equations hold:*

$$\min_{\mathcal{B} \subseteq \mathcal{A}_{\uparrow(v)}} P(\mathcal{A}_{\{v\}}|\mathcal{B}_{\uparrow+(v)}) = \min_{\mathcal{D} \in \mathcal{C}_{\uparrow(v)} \land \mathcal{D} \dot\in \mathcal{A}_{\uparrow(v)}} P(\mathcal{A}_{\{v\}}|\mathcal{D}) \quad (2)$$

$$\max_{\mathcal{B} \subseteq \mathcal{A}_{\uparrow(v)}} P(\mathcal{A}_{\{v\}}|\mathcal{B}_{\uparrow+(v)}) = \max_{\mathcal{D} \in \mathcal{C}_{\uparrow(v)} \land \mathcal{D} \dot\in \mathcal{A}_{\uparrow(v)}} P(\mathcal{A}_{\{v\}}|\mathcal{D}) \quad (3)$$

proof: We prove that the left-hand side of equation 2 (LHS) is less than or equal to the right-hand side of the equation (RHS) and vice versa. A similar argument proves equation 3.

To prove LHS $\leq$ RHS: we note that $\mathcal{B}$ ranges over all refinements of $\mathcal{A}_{\uparrow(v)}$. This includes the G-assignments where all the ancestors of $v$ are assigned sets of cardinality 1. For each of these cases, we have a unique assignment $\mathcal{F}$ that is complete w.r.t. the ancestors of $v$ such that $\mathcal{F} \dot\in \mathcal{B}$.

In Bayesian belief networks, a node is independent of any (indirect) ancestor given all of its parents, and thus, we have, for the above cases[5]:

$$P(\mathcal{A}_{\{v\}}|\mathcal{B}\uparrow^+ (v)) = P(\mathcal{A}_{\{v\}}|\mathcal{F})$$
$$= P(\mathcal{A}_{\{v\}}|\mathcal{F}_{\uparrow(v)}) \quad (4)$$

Now, since the RHS of equation 2 minimizes $P(\mathcal{A}_{\{v\}}|\mathcal{D})$ over complete assignments to the parents

---

[5]Actually, this is known to hold only for a *value* assigned to $v$, not for a set of values as here. However, since

$$P(\mathcal{A}_{\{v\}}|\mathcal{B}\uparrow^+ (v)) = \sum_{\mathcal{A}' \dot\in \mathcal{A}_{\{v\}} \land \mathcal{A}' \in \mathcal{C}_{\{v\}}} P(\mathcal{A}'|\mathcal{B}\uparrow^+ (v))$$

and the independence does hold for each $\mathcal{A}'$ (since $\mathcal{A}'$ it assigns exactly one value to $v$), then it also holds for the entire sum.

of $v$ that are included in $\mathcal{A}$, and for every such $\mathcal{D}$ there exists a G-assignment $\mathcal{B}$ that includes exactly one (complete w.r.t. the ancestors of $v$) assignment $\mathcal{F}$ that is a refinement of $\mathcal{D}$ such that equation 4 holds, then the LHS minimizes over a set that includes all the cases which are minimized over by the RHS, and thus we get LHS $\leq$ RHS.

To prove LHS $\geq$ RHS: let $\mathcal{B}$ be any G-assignment that is more refined than $\mathcal{A}$. Now, using conditioning we can write:

$$P(\mathcal{A}_{\{v\}}|\mathcal{B}_{\uparrow+(v)}) = \sum_{\mathcal{D} \dot\in \mathcal{B}_{\uparrow+(v)} \land \mathcal{D} \in \mathcal{C}_{\uparrow+(v)}} P(\mathcal{A}_{\{v\}}|\mathcal{D}) P(\mathcal{D}|\mathcal{B}_{\uparrow+(v)}) \quad (5)$$

But all $\mathcal{D}$ are disjoint, and range over all the complete assignments included in $\mathcal{B}_{\uparrow+(v)}$, and thus:

$$\sum_{\mathcal{D} \dot\in \mathcal{B}_{\uparrow+(v)} \land \mathcal{D} \in \mathcal{C}_{\uparrow+(v)}} P(\mathcal{D}|\mathcal{B}_{\uparrow+(v)}) = 1$$

Therefore, equation 5 is a convex sum, and we have:

$$\max_{\mathcal{D} \dot\in \mathcal{B}_{\uparrow+(v)} \land \mathcal{D} \in \mathcal{C}_{\uparrow+(v)}} P(\mathcal{A}_{\{v\}}|\mathcal{D}) \geq P(\mathcal{A}_{\{v\}}|\mathcal{B}_{\uparrow+(v)})$$

$$\min_{\mathcal{D} \dot\in \mathcal{B}_{\uparrow+(v)} \land \mathcal{D} \in \mathcal{C}_{\uparrow+(v)}} P(\mathcal{A}_{\{v\}}|\mathcal{D}) \leq P(\mathcal{A}_{\{v\}}|\mathcal{B}_{\uparrow+(v)}) \quad (6)$$

Since $\mathcal{D}$ is a complete assignment to exactly all the ancestors of $v$, then $v$ depends only on the assignment to its parents:

$$P(\mathcal{A}_{\{v\}}|\mathcal{D}) = P(\mathcal{A}_{\{v\}}|\mathcal{D}_{\uparrow(v)})$$

And thus minimizing (or maximizing) over all complete assignments to the ancestors of $v$ is equivalent to minimizing (or maximizing, respectively) over all complete assignments to the parents of $v$, and thus:

$$\max_{\mathcal{D} \dot\in \mathcal{B}_{\uparrow+(v)} \land \mathcal{D} \in \mathcal{C}_{\uparrow(v)}} P(\mathcal{A}_{\{v\}}|\mathcal{D}) \geq P(\mathcal{A}_{\{v\}}|\mathcal{B}_{\uparrow+(v)})$$

$$\min_{\mathcal{D} \dot\in \mathcal{B}_{\uparrow+(v)} \land \mathcal{D} \in \mathcal{C}_{\uparrow(v)}} P(\mathcal{A}_{\{v\}}|\mathcal{D}) \leq P(\mathcal{A}_{\{v\}}|\mathcal{B}_{\uparrow+(v)}) \quad (7)$$

Since $\mathcal{B}$ in equation 7 is an arbitrary refinement of $\mathcal{A}$, the equation holds for any such $\mathcal{B}$, in particular for the $\mathcal{B}$ that minimizes $P(\mathcal{A}_{\{v\}}|\mathcal{B}_{\uparrow+(v)})$. Now, this particular $\mathcal{B}$ is more refined than $\mathcal{A}$, and thus includes a (set-wise) smaller set of complete assignments to the parents of $v$ than does $\mathcal{A}$, and thus:

$$\min_{\mathcal{B} \subseteq \mathcal{A}_{\uparrow(v)}} P(\mathcal{A}_{\{v\}}|\mathcal{B}_{\uparrow+(v)}) \geq \min_{\mathcal{D} \dot\in \mathcal{B}_{\uparrow+(v)} \land \mathcal{D} \in \mathcal{C}_{\uparrow(v)}} P(\mathcal{A}_{\{v\}}|\mathcal{D})$$

$$\geq \min_{\mathcal{D} \in \mathcal{C}_{\uparrow(v)} \land \mathcal{D} \dot\in \mathcal{A}_{\uparrow(v)}} P(\mathcal{A}_{\{v\}}|\mathcal{D}) \quad (8)$$

Equation 2 follows. Equation 3 likewise follows from equation 7 ($\leq$), and from equation 4 ($\geq$), Q.E.D.

**Theorem 3** *Let $\mathcal{A}$ be a GIB assignment to a (positive distribution) Bayesian belief network. $P(\mathcal{A})$, the probability of $\mathcal{A}$ is the product:*

$$P(\mathcal{A}) = \prod_{v \in span(\mathcal{A})} P(\mathcal{A}_{\{v\}}|\mathcal{A}_{\uparrow(v)}) \qquad (9)$$

Proof outline: (complete proof omitted for lack of space). Assume, without loss of generality, that $\mathcal{A}$ assigns some value set to each and every node in the network. Let $B$, of cardinality $n$, be the set of nodes in the network. Define an integer index from 1 to $n$ on $B$ such that each node $v_i$ comes before all of its ancestors (where the subscript is the index). Clearly that is possible, as belief networks are directed acyclic graphs. Since the distribution is positive, it can be represented as a product of conditional probabilities, as follows:

$$P(\mathcal{A}) = \prod_{i=1}^{n} P(\mathcal{A}_{\{v_i\}}|\mathcal{A}_{\{v_j|n\geq j>i\}}) \qquad (10)$$

It is sufficient to prove that for every $n \geq i \geq 1$, the following equation holds:

$$P(\mathcal{A}_{\{v_i\}}|\mathcal{A}_{\{v_j|n\geq j>i\}}) = P(\mathcal{A}_{\{v_i\}}|\mathcal{A}_{\uparrow(v)}) \qquad (11)$$

We can separate out the nodes assigned by the conditioning term on the left-hand side of the above equation into parents of $v_i$, other ancestors of $v_i$, and all the rest. We then condition on all events that are included in $\mathcal{A}_{\uparrow(v)}$ (i.e. write $P(\mathcal{A}_{\{v_i\}}|\mathcal{A}_{\{v_j|n\geq j>i\}})$ as a sum of probability terms). Due to independence, we can drop some of the conditioning terms, and take some terms outside the summation, to get:

$$P(\mathcal{A}_{\{v_i\}}|\mathcal{A}_{\{v_j|n\geq j>i\}}) = P(\mathcal{A}_{\{v_i\}}|\mathcal{A}_{\uparrow(v)})\Sigma$$

where $\Sigma$ is a sum of conditional probabilities, which is shown to be equal to 1.

## References

[1] Eugene Charniak and Robert Goldman. A logic for semantic interpretation. In *Proceedings of the ACL Conference*, 1988.

[2] Eugene Charniak and Solomon E. Shimony. Probabilistic semantics for cost-based abduction. In *Proceedings of the 8th National Conference on AI*, August 1990.

[3] Gregory F. Cooper. The computational complexity of probabilistic inference using bayesian belief networks. *Artificial Intelligence*, 42 (2-3):393–405, 1990.

[4] Gregory F. Cooper and Herskovits. Edward. A bayesian method for the induction of probabilistic networks from data. Technical Report SMI-91-1, University of Pittsburgh, January 1991.

[5] Gregory Floyd Cooper. *NESTOR: A Computer-Based Medical Diagnosis Aid that Integrates Causal and Probabilistic Knowledge*. PhD thesis, Stanford University, 1984.

[6] Jerome A. Feldman and Yoram Yakimovsky. Decision theory and artificial intelligence: I. a semantics-based region analyzer. *Artificial Intelligence*, 5:349–371, 1974.

[7] Stuart Geeman and Donald Geeman. Stochastic relaxation, gibbs distributions and the bayesian restoration of images. *IEEE Transactions on Pattern Analysis and Machine Intelligence*, 6:721–741, 1984.

[8] Dan Geiger and David Heckerman. Advances in probabilistic reasoning. In *Proceedings of the 7th Conference on Uncertainty in AI*, 1991.

[9] Robert P. Goldman. *A Probabilistic Approach to Language Understanding*. PhD thesis, Brown University, 1990. Technical report CS-90-34.

[10] Jerry R. Hobbs, Mark Stickel, Paul Martin, and Douglas Edwards. Interpretation as abduction. In *Proceedings of the 26th Conference of the ACL*, 1988.

[11] Judea Pearl. *Probabilistic Reasoning in Intelligent Systems: Networks of Plausible Inference*. Morgan Kaufmann, San Mateo, CA, 1988.

[12] Judea Pearl and T. S. Verma. A theory of inferred causation. In *Knowledge Representation and Reasoning: Proceedings of the second International Conference*, pages 441–452, April 1991.

[13] Y. Peng and J. A. Reggia. A probabilistic causal model for diagnostic problem solving (parts 1 and 2). In *IEEE Transactions on Systems, Man and Cybernetics*, pages 146–162 and 395–406, 1987.

[14] R. D. Shachter. Evaluating influence diagrams. *Operations Research*, 34 (6):871–882, 1986.

[15] David B. Sher. Towards a normative theory of scientific evidence - a maximum likelihood solution. In *Proceedings of the 6th Conference on Uncertainty in AI*, pages 509–515, 1990.

[16] Solomon E. Shimony. Algorithms for finding irrelevance-based map assignments to belief networks. In *Proceedings of the 7th Conference on Uncertainty in AI*, 1991.

[17] Solomon E. Shimony. Explanation, irrelevance and statistical independence. In *AAAI Proceedings*, 1991.

[18] Solomon E. Shimony. *A Probabilistic Framework for Explanation*. PhD thesis, Brown University, 1991. Technical report CS-91-57.

[19] Mark E. Stickel. A prolog-like inference system for computing minimum-cost abductive explanations in natural-language interpretation. Technical Report 451, Artificial Intelligence Center, SRI, September 1988.